# Efficient Inference in Fully Connected CRFs with Gaussian Edge Potentials

**Philipp Krähenbühl**
Computer Science Department
Stanford University
philkr@cs.stanford.edu

**Vladlen Koltun**
Computer Science Department
Stanford University
vladlen@cs.stanford.edu

## Abstract

Most state-of-the-art techniques for multi-class image segmentation and labeling use conditional random fields defined over pixels or image regions. While region-level models often feature dense pairwise connectivity, pixel-level models are considerably larger and have only permitted sparse graph structures. In this paper, we consider fully connected CRF models defined on the complete set of pixels in an image. The resulting graphs have billions of edges, making traditional inference algorithms impractical. Our main contribution is a highly efficient approximate inference algorithm for fully connected CRF models in which the pairwise edge potentials are defined by a linear combination of Gaussian kernels. Our experiments demonstrate that dense connectivity at the pixel level substantially improves segmentation and labeling accuracy.

## 1 Introduction

Multi-class image segmentation and labeling is one of the most challenging and actively studied problems in computer vision. The goal is to label every pixel in the image with one of several predetermined object categories, thus concurrently performing recognition and segmentation of multiple object classes. A common approach is to pose this problem as maximum a posteriori (MAP) inference in a conditional random field (CRF) defined over pixels or image patches [8, 12, 18, 19, 9]. The CRF potentials incorporate smoothness terms that maximize label agreement between similar pixels, and can integrate more elaborate terms that model contextual relationships between object classes.

Basic CRF models are composed of unary potentials on individual pixels or image patches and pairwise potentials on neighboring pixels or patches [19, 23, 7, 5]. The resulting *adjacency CRF* structure is limited in its ability to model long-range connections within the image and generally results in excessive smoothing of object boundaries. In order to improve segmentation and labeling accuracy, researchers have expanded the basic CRF framework to incorporate hierarchical connectivity and higher-order potentials defined on image regions [8, 12, 9, 13]. However, the accuracy of these approaches is necessarily restricted by the accuracy of unsupervised image segmentation, which is used to compute the regions on which the model operates. This limits the ability of region-based approaches to produce accurate label assignments around complex object boundaries, although significant progress has been made [9, 13, 14].

In this paper, we explore a different model structure for accurate semantic segmentation and labeling. We use a *fully connected CRF* that establishes pairwise potentials on all pairs of pixels in the image. Fully connected CRFs have been used for semantic image labeling in the past [18, 22, 6, 17], but the complexity of inference in fully connected models has restricted their application to sets of hundreds of image regions or fewer. The segmentation accuracy achieved by these approaches is again limited by the unsupervised segmentation that produces the regions. In contrast, our model connects all

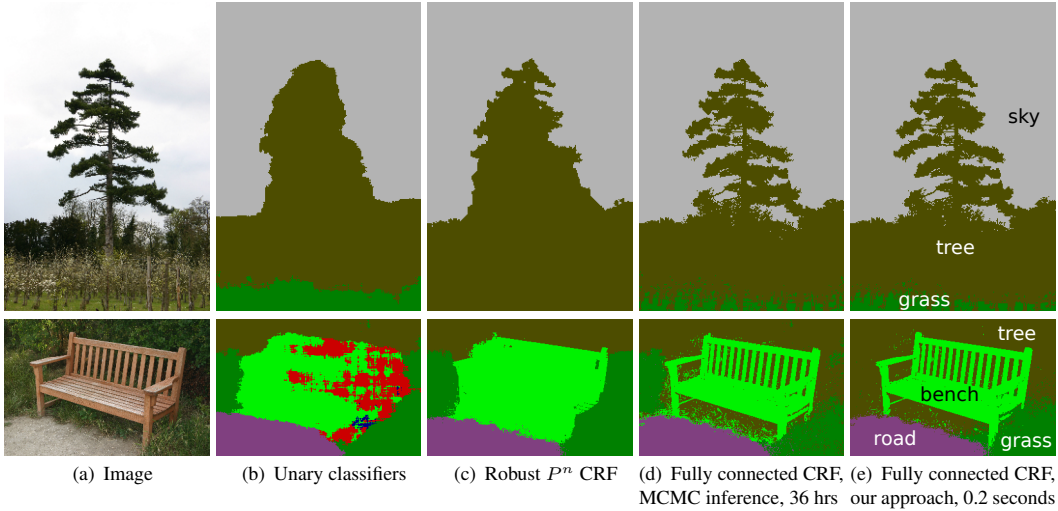

(a) Image  (b) Unary classifiers  (c) Robust $P^n$ CRF  (d) Fully connected CRF, MCMC inference, 36 hrs  (e) Fully connected CRF, our approach, 0.2 seconds pairs of individual pixels in the image, enabling greatly refined segmentation and labeling. The main challenge is the size of the model, which has tens of thousands of nodes and billions of edges even on low-resolution images.

Figure 1: Pixel-level classification with a fully connected CRF. (a) Input image from the MSRC-21 dataset. (b) The response of unary classifiers used by our models. (c) Classification produced by the Robust $P^n$ CRF [9]. (d) Classification produced by MCMC inference [17] in a fully connected pixel-level CRF model; the algorithm was run for 36 hours and only partially converged for the bottom image. (e) Classification produced by our inference algorithm in the fully connected model in 0.2 seconds.

Our main contribution is a highly efficient inference algorithm for fully connected CRF models in which the pairwise edge potentials are defined by a linear combination of Gaussian kernels in an arbitrary feature space. The algorithm is based on a mean field approximation to the CRF distribution. This approximation is iteratively optimized through a series of message passing steps, each of which updates a single variable by aggregating information from all other variables. We show that a mean field update of all variables in a fully connected CRF can be performed using Gaussian filtering in feature space. This allows us to reduce the computational complexity of message passing from quadratic to linear in the number of variables by employing efficient approximate high-dimensional filtering [16, 2, 1]. The resulting approximate inference algorithm is sublinear in the number of edges in the model.

Figure 1 demonstrates the benefits of the presented algorithm on two images from the MSRC-21 dataset for multi-class image segmentation and labeling. Figure 1(d) shows the results of approximate MCMC inference in fully connected CRFs on these images [17]. The MCMC procedure was run for 36 hours and only partially converged for the bottom image. We have also experimented with graph cut inference in the fully connected models [11], but it did not converge within 72 hours. In contrast, a single-threaded implementation of our algorithm produces a detailed pixel-level labeling in 0.2 seconds, as shown in Figure 1(e). A quantitative evaluation on the MSRC-21 and the PASCAL VOC 2010 datasets is provided in Section 6. To the best of our knowledge, we are the first to demonstrate efficient inference in fully connected CRF models at the pixel level.

## 2  The Fully Connected CRF Model

Consider a random field $\mathbf{X}$ defined over a set of variables $\{X_1, \ldots, X_N\}$. The domain of each variable is a set of labels $\mathcal{L} = \{l_1, l_2, \ldots, l_k\}$. Consider also a random field $\mathbf{I}$ defined over variables $\{I_1, \ldots, I_N\}$. In our setting, $\mathbf{I}$ ranges over possible input images of size $N$ and $\mathbf{X}$ ranges over possible pixel-level image labelings. $I_j$ is the color vector of pixel $j$ and $X_j$ is the label assigned to pixel $j$.

A conditional random field $(\mathbf{I}, \mathbf{X})$ is characterized by a Gibbs distribution $P(\mathbf{X}|\mathbf{I}) = \frac{1}{Z(\mathbf{I})} \exp(-\sum_{c \in \mathcal{C}_\mathcal{G}} \phi_c(\mathbf{X}_c|\mathbf{I}))$, where $\mathcal{G} = (\mathcal{V}, \mathcal{E})$ is a graph on $\mathbf{X}$ and each clique $c$

in a set of cliques $\mathcal{C}_\mathcal{G}$ in $\mathcal{G}$ induces a potential $\phi_c$ [15]. The Gibbs energy of a labeling $\mathbf{x} \in \mathcal{L}^N$ is $E(\mathbf{x}|\mathbf{I}) = \sum_{c \in \mathcal{C}_\mathcal{G}} \phi_c(\mathbf{x}_c|\mathbf{I})$. The maximum a posteriori (MAP) labeling of the random field is $\mathbf{x}^* = \arg\max_{\mathbf{x} \in \mathcal{L}^N} P(\mathbf{x}|\mathbf{I})$. For notational convenience we will omit the conditioning in the rest of the paper and use $\psi_c(\mathbf{x}_c)$ to denote $\phi_c(\mathbf{x}_c|\mathbf{I})$.

In the fully connected pairwise CRF model, $\mathcal{G}$ is the complete graph on $\mathbf{X}$ and $\mathcal{C}_\mathcal{G}$ is the set of all unary and pairwise cliques. The corresponding Gibbs energy is

$$E(\mathbf{x}) = \sum_i \psi_u(x_i) + \sum_{i<j} \psi_p(x_i, x_j), \qquad (1)$$

where $i$ and $j$ range from 1 to $N$. The unary potential $\psi_u(x_i)$ is computed independently for each pixel by a classifier that produces a distribution over the label assignment $x_i$ given image features. The unary potential used in our implementation incorporates shape, texture, location, and color descriptors and is described in Section 5. Since the output of the unary classifier for each pixel is produced independently from the outputs of the classifiers for other pixels, the MAP labeling produced by the unary classifiers alone is generally noisy and inconsistent, as shown in Figure 1(b).

The pairwise potentials in our model have the form

$$\psi_p(x_i, x_j) = \mu(x_i, x_j) \underbrace{\sum_{m=1}^{K} w^{(m)} k^{(m)}(\mathbf{f}_i, \mathbf{f}_j)}_{k(\mathbf{f}_i, \mathbf{f}_j)}, \qquad (2)$$

where each $k^{(m)}$ is a Gaussian kernel $k^{(m)}(\mathbf{f}_i, \mathbf{f}_j) = \exp(-\frac{1}{2}(\mathbf{f}_i - \mathbf{f}_j)^\mathrm{T} \Lambda^{(m)} (\mathbf{f}_i - \mathbf{f}_j))$, the vectors $\mathbf{f}_i$ and $\mathbf{f}_j$ are feature vectors for pixels $i$ and $j$ in an arbitrary feature space, $w^{(m)}$ are linear combination weights, and $\mu$ is a label compatibility function. Each kernel $k^{(m)}$ is characterized by a symmetric, positive-definite precision matrix $\Lambda^{(m)}$, which defines its shape.

For multi-class image segmentation and labeling we use contrast-sensitive two-kernel potentials, defined in terms of the color vectors $I_i$ and $I_j$ and positions $p_i$ and $p_j$:

$$k(\mathbf{f}_i, \mathbf{f}_j) = w^{(1)} \underbrace{\exp\left(-\frac{|p_i - p_j|^2}{2\theta_\alpha^2} - \frac{|I_i - I_j|^2}{2\theta_\beta^2}\right)}_{\text{appearance kernel}} + w^{(2)} \underbrace{\exp\left(-\frac{|p_i - p_j|^2}{2\theta_\gamma^2}\right)}_{\text{smoothness kernel}}. \qquad (3)$$

The *appearance kernel* is inspired by the observation that nearby pixels with similar color are likely to be in the same class. The degrees of nearness and similarity are controlled by parameters $\theta_\alpha$ and $\theta_\beta$. The *smoothness kernel* removes small isolated regions [19]. The parameters are learned from data, as described in Section 4.

A simple label compatibility function $\mu$ is given by the Potts model, $\mu(x_i, x_j) = [x_i \neq x_j]$. It introduces a penalty for nearby similar pixels that are assigned different labels. While this simple model works well in practice, it is insensitive to compatibility between labels. For example, it penalizes a pair of nearby pixels labeled "sky" and "bird" to the same extent as pixels labeled "sky" and "cat". We can instead learn a general symmetric compatibility function $\mu(x_i, x_j)$ that takes interactions between labels into account, as described in Section 4.

## 3 Efficient Inference in Fully Connected CRFs

Our algorithm is based on a mean field approximation to the CRF distribution. This approximation yields an iterative message passing algorithm for approximate inference. Our key observation is that message passing in the presented model can be performed using Gaussian filtering in feature space. This enables us to utilize highly efficient approximations for high-dimensional filtering, which reduce the complexity of message passing from quadratic to linear, resulting in an approximate inference algorithm for fully connected CRFs that is linear in the number of variables $N$ and sublinear in the number of edges in the model.

### 3.1 Mean Field Approximation

Instead of computing the exact distribution $P(\mathbf{X})$, the mean field approximation computes a distribution $Q(\mathbf{X})$ that minimizes the KL-divergence $\mathbf{D}(Q\|P)$ among all distributions $Q$ that can be expressed as a product of independent marginals, $Q(\mathbf{X}) = \prod_i Q_i(X_i)$ [10].

Minimizing the KL-divergence, while constraining $Q(\mathbf{X})$ and $Q_i(X_i)$ to be valid distributions, yields the following iterative update equation:

$$Q_i(x_i = l) = \frac{1}{Z_i} \exp \left\{ -\psi_u(x_i) - \sum_{l' \in \mathcal{L}} \mu(l, l') \sum_{m=1}^{K} w^{(m)} \sum_{j \neq i} k^{(m)}(\mathbf{f}_i, \mathbf{f}_j) Q_j(l') \right\}. \quad (4)$$

A detailed derivation of Equation 4 is given in the supplementary material. This update equation leads to the following inference algorithm:

---
**Algorithm 1** Mean field in fully connected CRFs

Initialize $Q$      ▷ $Q_i(x_i) \leftarrow \frac{1}{Z_i} \exp\{-\phi_u(x_i)\}$
**while** not converged **do**      ▷ See Section 6 for convergence analysis
     $\tilde{Q}_i^{(m)}(l) \leftarrow \sum_{j \neq i} k^{(m)}(\mathbf{f}_i, \mathbf{f}_j) Q_j(l)$ for all $m$      ▷ **Message passing** from all $X_j$ to all $X_i$
     $\hat{Q}_i(x_i) \leftarrow \sum_{l \in \mathcal{L}} \mu^{(m)}(x_i, l) \sum_m w^{(m)} \tilde{Q}_i^{(m)}(l)$      ▷ **Compatibility transform**
     $Q_i(x_i) \leftarrow \exp\{-\psi_u(x_i) - \hat{Q}_i(x_i)\}$      ▷ **Local update**
     normalize $Q_i(x_i)$
**end while**

---

Each iteration of Algorithm 1 performs a message passing step, a compatibility transform, and a local update. Both the compatibility transform and the local update run in linear time and are highly efficient. The computational bottleneck is message passing. For each variable, this step requires evaluating a sum over all other variables. A naive implementation thus has quadratic complexity in the number of variables $N$. Next, we show how approximate high-dimensional filtering can be used to reduce the computational cost of message passing to linear.

### 3.2 Efficient Message Passing Using High-Dimensional Filtering

From a signal processing standpoint, the message passing step can be expressed as a convolution with a Gaussian kernel $G_{\Lambda^{(m)}}$ in feature space:

$$\tilde{Q}_i^{(m)}(l) = \underbrace{\sum_{j \in \mathcal{V}} k^{(m)}(\mathbf{f}_i, \mathbf{f}_j) Q_j(l) - Q_i(l)}_{\text{message passing}} = \underbrace{[G_{\Lambda^{(m)}} \otimes Q(l)](\mathbf{f}_i)}_{\overline{Q}_i^{(m)}(l)} - Q_i(l). \quad (5)$$

We subtract $Q_i(l)$ from the convolved function $\overline{Q}_i^{(m)}(l)$ because the convolution sums over all variables, while message passing does not sum over $Q_i$.

This convolution performs a low-pass filter, essentially band-limiting $\overline{Q}_i^{(m)}(l)$. By the sampling theorem, this function can be reconstructed from a set of samples whose spacing is proportional to the standard deviation of the filter [20]. We can thus perform the convolution by downsampling $Q(l)$, convolving the samples with $G_{\Lambda^{(m)}}$, and upsampling the result at the feature points [16].

---
**Algorithm 2** Efficient message passing: $\overline{Q}_i^{(m)}(l) = \sum_{j \in \mathcal{V}} k^{(m)}(\mathbf{f}_i, \mathbf{f}_j) Q_j(l)$

$Q_\downarrow(l) \leftarrow \text{downsample}(Q(l))$      ▷ **Downsample**
$\forall_{i \in \mathcal{V}_\downarrow} \overline{Q}_{\downarrow i}^{(m)}(l) \leftarrow \sum_{j \in \mathcal{V}_\downarrow} k^{(m)}(\mathbf{f}_{\downarrow i}, \mathbf{f}_{\downarrow j}) Q_{\downarrow j}(l)$      ▷ **Convolution** on samples $\mathbf{f}_\downarrow$
$\overline{Q}^{(m)}(l) \leftarrow \text{upsample}(\overline{Q}_\downarrow^{(m)}(l))$      ▷ **Upsample**

---

A common approximation to the Gaussian kernel is a truncated Gaussian, where all values beyond two standard deviations are set to zero. Since the spacing of the samples is proportional to the standard deviation, the support of the truncated kernel contains only a constant number of sample points. Thus the convolution can be approximately computed at each sample by aggregating values from only a constant number of neighboring samples. This implies that approximate message passing can be performed in $O(N)$ time [16].

High-dimensional filtering algorithms that follow this approach can still have computational complexity exponential in $d$. However, a clever filtering scheme can reduce the complexity of the convolution operation to $O(Nd)$. We use the permutohedral lattice, a highly efficient convolution data structure that tiles the feature space with simplices arranged along $d+1$ axes [1]. The permutohedral lattice exploits the separability of unit variance Gaussian kernels. Thus we need to apply a whitening transform $\tilde{\mathbf{f}} = U\mathbf{f}$ to the feature space in order to use it. The whitening transformation is found using the Cholesky decomposition of $\Lambda^{(m)}$ into $UU^{\mathrm{T}}$. In the transformed space, the high-dimensional convolution can be separated into a sequence of one-dimensional convolutions along the axes of the lattice. The resulting approximate message passing procedure is highly efficient even with a fully sequential implementation that does not make use of parallelism or the streaming capabilities of graphics hardware, which can provide further acceleration if desired.

## 4 Learning

We learn the parameters of the model by piecewise training. First, the boosted unary classifiers are trained using the JointBoost algorithm [21], using the features described in Section 5. Next we learn the appearance kernel parameters $w^{(1)}$, $\theta_\alpha$, and $\theta_\beta$ for the Potts model. $w^{(1)}$ can be found efficiently by a combination of expectation maximization and high-dimensional filtering. Unfortunately, the kernel widths $\theta_\alpha$ and $\theta_\beta$ cannot be computed effectively with this approach, since their gradient involves a sum of non-Gaussian kernels, which are not amenable to the same acceleration techniques. We found it to be more efficient to use grid search on a holdout validation set for all three kernel parameters $w^{(1)}$, $\theta_\alpha$ and $\theta_\beta$.

The smoothness kernel parameters $w^{(2)}$ and $\theta_\gamma$ do not significantly affect classification accuracy, but yield a small visual improvement. We found $w^{(2)} = \theta_\gamma = 1$ to work well in practice.

The compatibility parameters $\mu(a, b) = \mu(b, a)$ are learned using L-BFGS to maximize the log-likelihood $\ell(\mu : \mathcal{I}, \mathcal{T})$ of the model for a validation set of images $\mathcal{I}$ with corresponding ground truth labelings $\mathcal{T}$. L-BFGS requires the computation of the gradient of $\ell$, which is intractable to estimate exactly, since it requires computing the gradient of the partition function $Z$. Instead, we use the mean field approximation described in Section 3 to estimate the gradient of $Z$. This leads to a simple approximation of the gradient for each training image:

$$\frac{\partial}{\partial \mu(a,b)} \ell(\mu : \mathcal{I}^{(n)}, \mathcal{T}^{(n)}) \approx -\sum_i \mathcal{T}_i^{(n)}(a) \sum_{j \neq i} k(\mathbf{f}_i, \mathbf{f}_j) \mathcal{T}_j^{(n)}(b) + \sum_i Q_i(a) \sum_{j \neq i} k(\mathbf{f}_i, \mathbf{f}_j) Q_i(b), \quad (6)$$

where $(\mathcal{I}^{(n)}, \mathcal{T}^{(n)})$ is a single training image with its ground truth labeling and $\mathcal{T}^{(n)}(a)$ is a binary image in which the $i$th pixel $\mathcal{T}_i^{(n)}(a)$ has value 1 if the ground truth label at the $i$th pixel of $\mathcal{T}^{(n)}$ is $a$ and 0 otherwise. A detailed derivation of Equation 6 is given in the supplementary material.

The sums $\sum_{j \neq i} k(\mathbf{f}_i, \mathbf{f}_j) \mathcal{T}_j(b)$ and $\sum_{j \neq i} k(\mathbf{f}_i, \mathbf{f}_j) Q_i(b)$ are both computationally expensive to evaluate directly. As in Section 3.2, we use high-dimensional filtering to compute both sums efficiently. The runtime of the final learning algorithm is linear in the number of variables $N$.

## 5 Implementation

The unary potentials used in our implementation are derived from TextonBoost [19, 13]. We use the 17-dimensional filter bank suggested by Shotton et al. [19], and follow Ladický et al. [13] by adding color, histogram of oriented gradients (HOG), and pixel location features. Our evaluation on the MSRC-21 dataset uses this extended version of TextonBoost for the unary potentials. For the VOC 2010 dataset we include the response of bounding box object detectors [4] for each object class as 20 additional features. This increases the performance of the unary classifiers on the VOC 2010 from 13% to 22%. We gain an additional 5% by training a logistic regression classifier on the responses of the boosted classifier.

For efficient high-dimensional filtering, we use a publicly available implementation of the permutohedral lattice [1]. We found a downsampling rate of one standard deviation to work best for all our experiments. Sampling-based filtering algorithms underestimate the edge strength $k(\mathbf{f}_i, \mathbf{f}_j)$ for very similar feature points. Proper normalization can cancel out most of this error. The permutohedral lattice allows for two types of normalizations. A global normalization by the average kernel strength $\hat{k} = \frac{1}{N} \sum_{i,j} k(\mathbf{f}_i, \mathbf{f}_j)$ can correct for constant error. A pixelwise normalization by $\hat{k}_i = \sum_j k(\mathbf{f}_i, \mathbf{f}_j)$ handles regional errors as well, but violates the CRF symmetry assumption $\psi_p(x_i, x_j) = \psi_p(x_j, x_i)$. We found the pixelwise normalization to work better in practice.

## 6  Evaluation

We evaluate the presented algorithm on two standard benchmarks for multi-class image segmentation and labeling. The first is the MSRC-21 dataset, which consists of 591 color images of size $320 \times 213$ with corresponding ground truth labelings of 21 object classes [19]. The second is the PASCAL VOC 2010 dataset, which contains 1928 color images of size approximately $500 \times 400$, with a total of 20 object classes and one background class [3]. The presented approach was evaluated alongside the adjacency (grid) CRF of Shotton et al. [19] and the Robust $P^n$ CRF of Kohli et al. [9], using publicly available reference implementations. To ensure a fair comparison, all models used the unary potentials described in Section 5. All experiments were conducted on an Intel i7-930 processor clocked at 2.80GHz. Eight CPU cores were used for training; all other experiments were performed on a single core. The inference algorithm was implemented in a single CPU thread.

**Convergence.** We first evaluate the convergence of the mean field approximation by analyzing the KL-divergence between $Q$ and $P$. Figure 2 shows the KL-divergence between $Q$ and $P$ over successive iterations of the inference algorithm. The KL-divergence was estimated up to a constant as described in supplementary material. Results are shown for different standard deviations $\theta_\alpha$ and $\theta_\beta$ of the kernels. The graphs were aligned at 20 iterations for visual comparison. The number of iterations was set to 10 in all subsequent experiments.

**MSRC-21 dataset.** We use the standard split of the dataset into $45\%$ training, $10\%$ validation and $45\%$ test images [19]. The unary potentials were learned on the training set, while the parameters of all CRF models were learned using holdout validation. The total CRF training time was 40 minutes. The learned label compatibility function performed on par with the Potts model on this dataset. Figure 3 provides qualitative and quantitative results on the dataset. We report the standard measures of multi-class segmentation accuracy: "global" denotes the overall percentage of correctly classified image pixels and "average" is the unweighted average of per-category classification accuracy [19, 9]. The presented inference algorithm on the fully connected CRF significantly outperforms the other models, evaluated against the standard ground truth data provided with the dataset.

The ground truth labelings provided with the MSRC-21 dataset are quite imprecise. In particular, regions around object boundaries are often left unlabeled. This makes it difficult to quantitatively evaluate the performance of algorithms that strive for pixel-level accuracy. Following Kohli et al. [9], we manually produced accurate segmentations and labelings for a set of images from the MSRC-21 dataset. Each image was fully annotated at the pixel level, with careful labeling around complex boundaries. This labeling was performed by hand for 94 representative images from the MSRC-21 dataset. Labeling a single image took 30 minutes on average. A number of images from this "accurate ground truth" set are shown in Figure 3. Figure 3 reports segmentation accuracy against this ground truth data alongside the evaluation against the standard ground truth. The results were obtained using 5-fold cross validation, where $\frac{4}{5}$ of the 94 images were used to train the CRF pa-

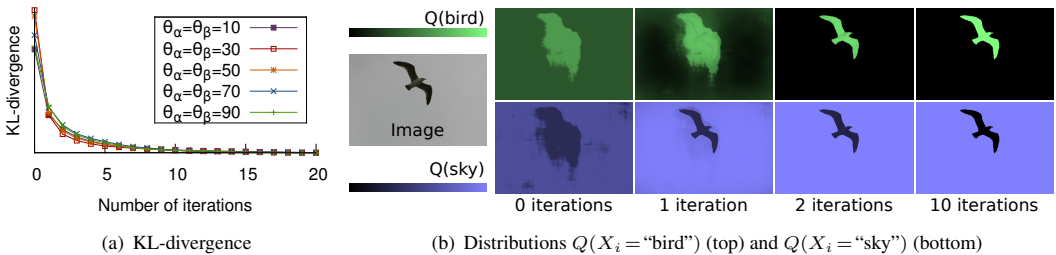

(a) KL-divergence    (b) Distributions $Q(X_i = \text{"bird"})$ (top) and $Q(X_i = \text{"sky"})$ (bottom)

Figure 2: Convergence analysis. (a) KL-divergence of the mean field approximation during successive iterations of the inference algorithm, averaged across 94 images from the MSRC-21 dataset. (b) Visualization of convergence on distributions for two class labels over an image from the dataset.

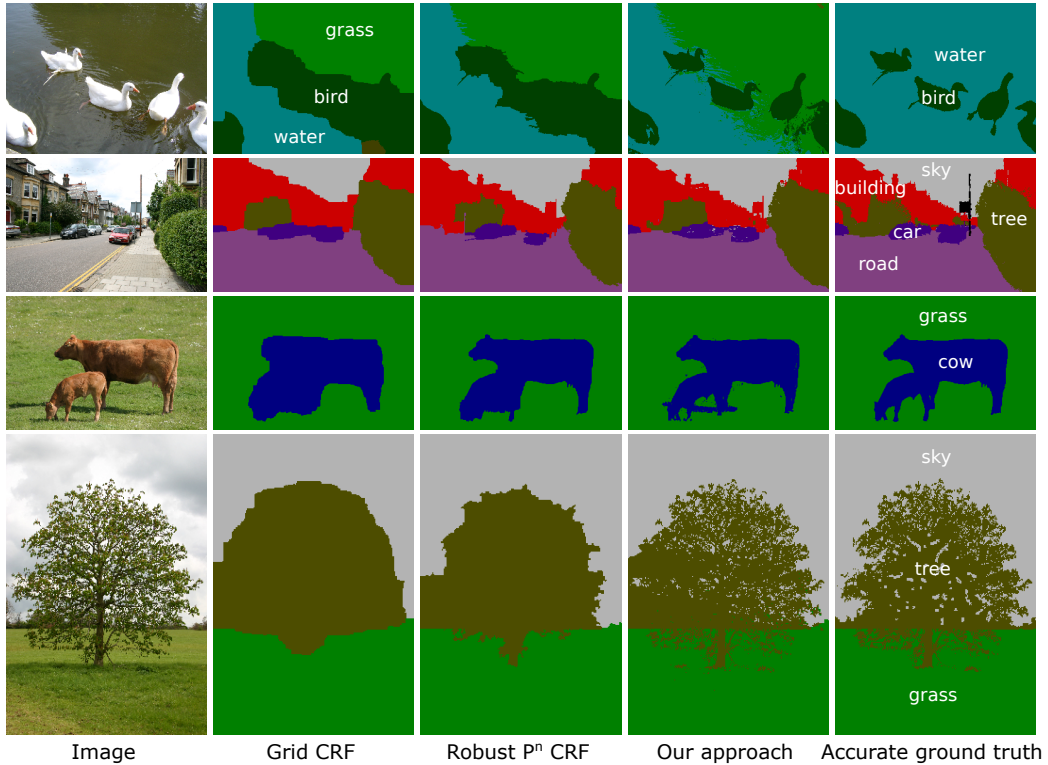

Figure 3: Qualitative and quantitative results on the MSRC-21 dataset.

|  | Runtime | Standard ground truth | | Accurate ground truth | |
| --- | --- | --- | --- | --- | --- |
|  |  | Global | Average | Global | Average |
| Unary classifiers | − | 84.0 | 76.6 | 83.2 ± 1.5 | 80.6 ± 2.3 |
| Grid CRF | 1s | 84.6 | 77.2 | 84.8 ± 1.5 | 82.4 ± 1.8 |
| Robust $P^n$ CRF | 30s | 84.9 | 77.5 | 86.5 ± 1.0 | 83.1 ± 1.5 |
| Fully connected CRF | **0.2s** | **86.0** | **78.3** | **88.2** ± 0.7 | **84.7** ± 0.7 |

rameters. The unary potentials were learned on a separate training set that did not include the 94 accurately annotated images.

We also adopt the methodology proposed by Kohli et al. [9] for evaluating segmentation accuracy around boundaries. Specifically, we count the relative number of misclassified pixels within a narrow band ("trimap") surrounding actual object boundaries, obtained from the accurate ground truth images. As shown in Figure 4, our algorithm outperforms previous work across all trimap widths.

**PASCAL VOC 2010.** Due to the lack of a publicly available ground truth labeling for the test set in the PASCAL VOC 2010, we use the training and validation data for all our experiments. We randomly partitioned the images into 3 groups: 40% training, 15% validation, and 45% test set. Segmentation accuracy was measured using the standard VOC measure [3]. The unary potentials were learned on the training set and yielded an average classification accuracy of 27.6%. The parameters for the Potts potentials in the fully connected CRF model were learned on the validation set. The fully connected model with Potts potentials yielded an average classification accuracy of 29.1%. The label compatibility function, learned on the validation set, further increased the classification accuracy to 30.2%. For comparison, the grid CRF achieves 28.3%. Training time was 2.5 hours and inference time is 0.5 seconds. Qualitative results are provided in Figure 5.

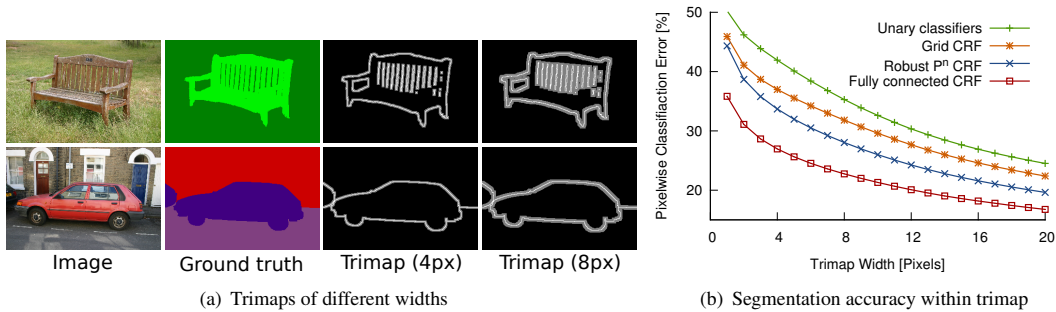

(a) Trimaps of different widths      (b) Segmentation accuracy within trimap

Figure 4: Segmentation accuracy around object boundaries. (a) Visualization of the "trimap" measure. (b) Percent of misclassified pixels within trimaps of different widths.

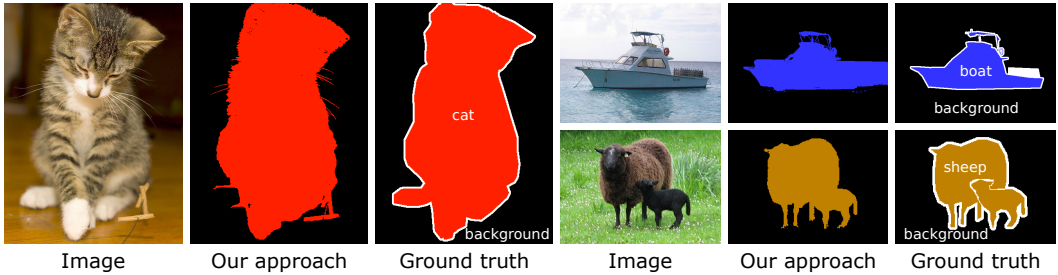

Figure 5: Qualitative results on the PASCAL VOC 2010 dataset. Average segmentation accuracy was 30.2%.

**Long-range connections.** We have examined the value of long-range connections in our model by varying the spatial and color ranges $\theta_\alpha$ and $\theta_\beta$ of the appearance kernel and analyzing the resulting classification accuracy. For this experiment, $w^{(1)}$ was held constant and $w^{(2)}$ was set to 0. The results are shown in Figure 6. Accuracy steadily increases as longer-range connections are added, peaking at spatial standard deviation of $\theta_\alpha = 61$ pixels and color standard deviation $\theta_\beta = 11$. At this setting, more than 50% of the pairwise potential energy in the model was assigned to edges of length 35 pixels or higher. However, long-range connections can also propagate misleading information, as shown in Figure 7.

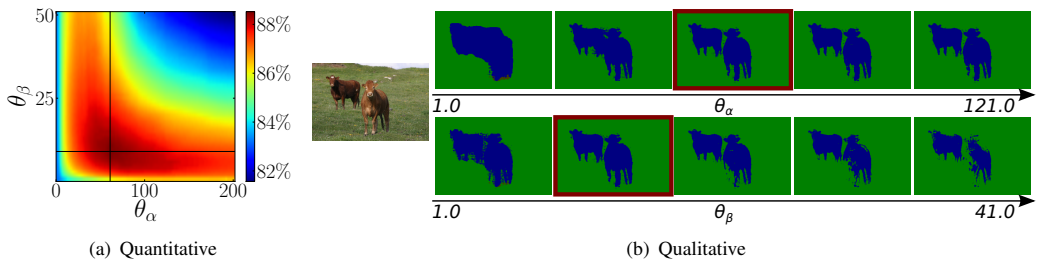

(a) Quantitative   (b) Qualitative

Figure 6: Influence of long-range connections on classification accuracy. (a) Global classification accuracy on the 94 MSRC images with accurate ground truth, as a function of kernel parameters $\theta_\alpha$ and $\theta_\beta$. (b) Results for one image across two slices in parameter space, shown as black lines in (a).

**Discussion.** We have presented a highly efficient approximate inference algorithm for fully connected CRF models. Our results demonstrate that dense pixel-level connectivity leads to significantly more accurate pixel-level classification performance. Our single-threaded implementation processes benchmark images in a fraction of a second and the algorithm can be parallelized for further performance gains.

**Acknowledgements.** Philipp Krähenbühl was supported in part by a Stanford Graduate Fellowship. We are grateful to Daphne Koller, Andrew Adams and Jongmin Baek for helpful discussions. Sergey Levine and Vangelis Kalogerakis provided comments on a draft of this paper.

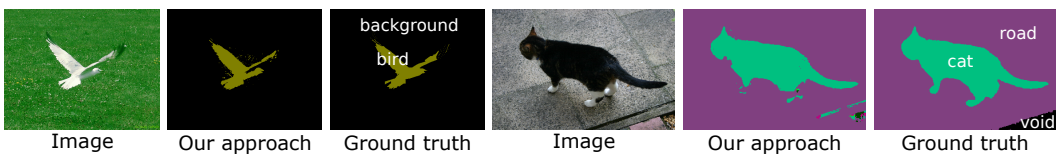

Figure 7: Failure cases on images from the PASCAL VOC 2010 (left) and the MSRC-21 (right). Long-range connections propagated misleading information, eroding the bird wing in the left image and corrupting the legs of the cat on the right.